%% file: paper.tex
\RequirePackage{fix-cm}
\documentclass[twocolumn]{svjour3}          % twocolumn
\smartqed  % flush right qed marks, e.g. at end of proof
\usepackage{graphicx}
\usepackage{natbib}

\usepackage{soul}
\usepackage{url}
\usepackage[utf8]{inputenc}
\usepackage[small]{caption}
\usepackage{subcaption}
\usepackage{graphicx}
\usepackage{amsmath}
\usepackage{booktabs}
\usepackage{algorithm}
\usepackage{algorithmic}
\usepackage{epsfig}
\usepackage{amssymb}
\usepackage{color}
\usepackage[table]{xcolor}
\usepackage{hhline}
\usepackage{stfloats}
\definecolor{Gray}{gray}{0.85}
\usepackage{diagbox}
\usepackage{multirow}

\newcommand{\eqnref}[1]{Eq.~(\ref{eqn:#1})}
\newcommand{\figref}[1]{Fig.~\ref{fig:#1}}
\newcommand{\tabref}[1]{Table~\ref{tbl:#1}}

\usepackage{dsfont}
\usepackage{pifont} % http://ctan.org/pkg/pifont

\usepackage[pagebackref=true,breaklinks=true,colorlinks=True,bookmarks=false,linkcolor=blue,citecolor=blue]{hyperref}

\input{figs}
\input{tabs}

\begin{document}
\sloppy
\title{Dynamical Deep Generative Latent Modeling of 3D Skeletal Motion%\thanks{Grants or other notes
}


\author{Amirreza Farnoosh    \and
        Sarah Ostadabbas %etc.
}


\institute{A. Farnoosh and S. Ostadabbas \at
              Augmented Cognition Lab, Electrical and Computer Engineering Department, Northeastern University.
              %\email{shuliu@ece.neu.edu ,sehgal.n@husky.neu.edu,ostadabbas@ece.neu.edu}           %  \\
}

\date{Received: date / Accepted: date}

\maketitle

\begin{abstract}
%\com{Papers will consist of maximum 6 pages (text, figures and tables) +1 page for references only.}
In this paper, we propose a Bayesian switching dynamical model for segmentation of 3D pose data over time that uncovers interpretable patterns in the data and is generative. Our model decomposes highly correlated skeleton data into a set of few spatial basis of switching temporal processes in a low-dimensional latent framework. We parameterize these temporal processes with regard to a switching deep vector autoregressive prior in order to accommodate both multimodal and higher-order nonlinear inter-dependencies. This results in a dynamical deep generative latent model that parses the meaningful intrinsic states in the dynamics of 3D pose data using approximate variational inference, and enables a realistic low-level dynamical generation and segmentation of complex skeleton movements. Our experiments on four biological motion data containing bat flight, salsa dance, walking, and golf datasets substantiate superior performance of our model in comparison with the state-of-the-art methods\footnote{The source code, datasets, and rendered 3D models are provided as supplementary materials.}.
\end{abstract}

\section{Introduction}
Analyzing 3D motion capture datasets illustrating dynamical motions of a subject is the key processing step in many applications, including highlighting movement patterns of an athlete to optimize their performance, probing behavior of an endangered animal, and monitoring mobility of a patient in a rehabilitation study, to name a few \citep{mot}. In all these applications, body pose data contained in the motion capture sequence describes temporal evolution of specific phenomena or tasks and is switching between potentially limited number of states each representing a specific regime.

The efforts in quantifying complex kinematics of biological mechanisms in a lower dimensional subspace has led to the successful design of bio-inspired robots that can mimic their
biological counterparts to a great extent \citep{birch2000design,gong2016kinematic,hoff2016synergistic}. Many such works have used the statistical methods such as principal component analysis  (PCA) \citep{jolliffe1986principal}, for dimensionality reduction. 
For instance, \cite{santello1998postural} showed that 80\% of variance of grasping motion in humans can be described by its first two postural synergies.  \cite{riskin2008quantifying} discovered that approximating motion of bat wing with only one third of its principal components accounted for 95\% of variance of the articulated skeleton. 
%For instance, it is shown that greater than 80\% of the variance of static grasping data in humans could be described by the first two postural synergies, i.e. principal components \cite{santello1998postural}. Also, \cite{riskin2008quantifying} found that there are three groupings of joints in a bat wing that move together, accounting for 70\% of the joints. This study also discovered that approximating the bat’s motion with only one third of the principal components accounted for 95\% of the variance of the articulated skeleton. 

Motion data segmentation, on the other hand, has been extensively studied in the context of probabilistic PCA and Gaussian mixture model (GMM) \citep{barbivc2004segmenting}, kernelized temporal cut \citep{gong2012kernelized,gong2013structured}, sparse subspace clustering \citep{elhamifar2015dissimilarity,xia2017human,zhou2020multi,xia2020nonconvex}, kernel k-means and spectral clustering \citep{zhou2012hierarchical}, neighborhood graph \citep{kruger2016efficient,wu2017novel,chen2018automatic}, dynamic time warping \citep{papoutsakis2017temporal}, and topic modeling \citep{patrona2018motion}. However, these models are specifically designed for temporal clustering, and inherently do not model temporal dynamics, therefore are not generative.

The strong spatio-temporal correlation among joints of a human's or animal's skeleton captured by 3D motion capture data as well as clear sparseness in these types of data motivate the utilization of probabilistic models that can learn  underlying interpretable states from data and extract its low-dimensional motion patterns. 
%Specifically, in this paper, we focus on the application of bat flight modeling, with the aim of modeling how bats flight over time in a nonlinear dynamical system setting with an interpretable low-dimensional representation. An illustration of the bat skeleton and its joint structure is provided in \figref{batjoints}. To this end, we benefit from the properties of hidden Markov models (HMMs), \emph{nonlinear} vector auto-regressive (VAR) models, state-space models and deep learning all in a unified model. 
%\batjointsfig 
%\par
%To this end, we introduce a Bayesian switching dynamical model for segmentation of 3D pose data over time that uncovers interpretable patterns in the data and is generative. Specifically, we use a low-dimensional latent variable model to decompose a highly correlated skeleton data into a set of few spatial basis of switching temporal processes. These temporal processes are in turn parameterized with regard to a switching deep vector autoregressive (VAR) prior in order to accommodate both multimodal and higher-order nonlinear inter-dependencies. This results in a dynamical deep generative latent model that parses the meaningful intrinsic states in the dynamics of 3D pose data using approximate variational inference, and enables a realistic low-level dynamical generation (and segmentation) of complex skeleton movements. 
To this end, we propose a Bayesian state switching model for dynamical segmentation of 3D pose data that uncovers interpretable motion patterns and is generative. Specifically, we employ a low-dimensional deep generative latent model to decompose highly correlated skeleton data into a set of few spatial basis of switching auto-regressive temporal processes. This results in a flexible model that accommodates multimodal and higher-order nonlinear inter-dependencies, parses meaningful temporal modes in 3D pose data, and enables low-level dynamical generation and segmentation of complex skeleton movements. 
We demonstrate superior performance of our model on four biological motion data including bat flight, human salsa dance, walking and golf datasets in terms of learning interpretable states, prediction accuracy, and dynamical generation.
%We demonstrate the performance of our model in learning of interpretable states/regime for bat flight as well as human salsa dance data. Our results substantiate the superior performance of the proposed model in comparison with the state-of-the-art methods.

\graphicalmodel
\section{Related Works}
%\subsection{Dynamical systems modeling}
\textbf{Dynamical systems modeling --} 
Switching linear dynamical systems (SLDS) have long been investigated in the literature, \cite{ackerson1970state,chang1978state,hamilton1990analysis,ghahramani1996switching,murphy1998switching,fox2009nonparametric}. These models decompose time series data into series of simpler, repeated dynamical modes represented by discrete and continuous latent states. In SLDS framework, the transitions between discrete states are independent of their associated continuous values. This problem is addressed in recurrent switching linear dynamical system (rSLDS) \citep{linderman2017bayesian,nassar2019tree, becker2019switching}, by allowing the discrete state transition probabilities to depend on their preceding continuous states.
The rSLDS dynamical capacity is however limited as it assumes first-order linear dynamics. A recent work, \citep{farnoosh2020deepp}, extends these models by adopting nonlinear and higher-order multimodal dependencies through a deep switching autoregressive framework. We build our model on top of this framework and customize that for generative segmentation of motion data. This makes our model flexible for complex auto-regressive relations in motion sequences.

From another perspective, dynamical matrix factorization is used in \cite{sun2014collaborative,cai2015facets,bahadori2014fast,yu2016temporal,takeuchi2017autoregressive} for modeling linear dynamics in their low-dimensional temporal factors. Several studies have also employed neural networks for non-linear state-space modeling \citep{watter2015embed,karl2017deep,krishnan2017structured,fraccaro2017disentangled,becker2019recurrent,farnoosh2020deep}, which are restricted to first-order Markovian dependencies, and for time series prediction, \citep{chang2018memory,lai2018modeling,rangapuram2018deep,li2019enhancing,sen2019think,salinas2020deepar}, which most of them are non-probabilistic. 
%They estimate model parameters from input data using recurrent neural networks (RNNs), e.g., in \cite{chang2018memory,lai2018modeling,salinas2020deepar}, Transformers, in \cite{li2019enhancing}, or temporal convolution networks (TCNs), in \cite{sen2019think}.

%\vspace{-.45cm}
%\subsection{Motion segmentation}
\noindent\textbf{Motion segmentation --}
Segmentation of 3D motion capture data has been studied for many years. Here, we review recent works in this field. \cite{kruger2016efficient} introduced an automated method for temporal segmentation of human motion data into distinct actions based on a self-similarity matrix extracted from motion sequences of subsequent data points over a window. 
\cite{papoutsakis2017temporal} used dynamic time warping (DTW) for co-segmenting all pairs of motion sub-sequences that represent the same action in an unsupervised manner.
\cite{wu2017novel} used a combination of normalized cut model and weighted kernel k-means (NCWKK) for behavior segmentation of human motion capture data in its high-dimensional space.
\cite{xia2017human} used a sparse subspace clustering framework with geodesic exponential kernel and multi-view reconstruction for motion data segmentation that is able to model their underlying Riemannian manifold and is robust to non-Gaussian noise. \cite{chen2018automatic} developed a data structure called segment-graph by leveraging information bottleneck method
with minimum description length principle to temporally cluster data values via an average-longest-path optimization on this segment-graph.
\cite{patrona2018motion} performed online action detection and recognition based on an efficient linear search (ELS) approach proposed in \cite{meshry2016linear}. For this, local features, called gesturelets, capturing both skeleton and kinematics information, are extracted from 3D skeleton joint positions and are subsequently clustered into a bag-of-gesturelets (BoG) model. Then, binary linear classifiers are trained to identify decision boundaries for each specific action.
\cite{zhou2020multi} proposed a  multi level transfer subspace learning framework for human motion segmentation. To capture multi-level
structural information, their model factorize labeled source data, from some related task, and target task data into multi-layer feature spaces based on a deep non-negative matrix factorization (NMF) model. Finally, an affinity matrix is constructed and segmentation results are obtained by using the normalized cuts algorithm. \cite{xia2020nonconvex} proposed an unsupervised low-rank sparse subspace learning framework with non-convex relaxation for key-frame extraction and segmentation of motion capture data. 
%However, these works do not explicitly model temporal dynamics in motion data, and are not generative.

Although the aforementioned works are able to cluster motion sequences, they do not explicitly model temporal dynamics, and are not generative.
An exception is the work of \cite{nakamura2017segmenting} which proposed a Gaussian process-hidden semi-Markov model (GP-HSMM) for unsupervised motion segmentation. However, linear dynamical assumptions in this model fall short of capturing complex temporal dependencies in motion data.

%\noindent\textbf{Our Contributions --} In contrast to previous works which merely cluster motion sequences and are not generative, our method, sketched in \figref{graph}, 
%\begin{itemize}
%    \item segments motion data from a dynamical perspective by explicitly modeling temporal dynamics,
%    \item hence, allows low-level dynamical generation of complex skeleton movements. 
%    \item Specifically, it welcomes multimodal, higher-order, and nonlinear temporal relations in motion data by employing a deep switching auto-regressive latent model.
%\end{itemize}
%We focus on bat flight and human salsa dance, one being important in bio-inspired robotic design and the other in activity training as well as human behavior understanding.

\noindent\textbf{Our Contributions --} In contrast to previous works which merely cluster motion sequences and are not generative, our method, sketched in \figref{graph}, (i) segments motion data from a dynamical perspective by explicitly modeling temporal dynamics, hence, (ii) allows dynamical generation of skeletal movements from low-level representations. Specifically, (iii) it welcomes multimodal, higher-order, and nonlinear temporal relations in motion data by employing a deep switching auto-regressive latent model. We focus on four biological motion data including bat flight, human salsa dance, walking, and golf datasets, the first being important in bio-inspired robotic design and the others in activity training as well as human behavior understanding.
%%We focus on bat flight and human salsa dance, one being important in bio-inspired robotic design and the other in activity training as well as human behavior understanding.

%We will explain the detailed formulation of our proposed model in the next section.
%for unsupervised time series segmentation

\section{Problem Formulation}
\label{sec:prob}
We consider a set of $N$ motion datasets $\{X_1, \dots, X_N\}$, where each $X_n \in \mathbb{R}^{T \times D}$ contains $T$ time points and $D$ spatial coordinates (e.g., $D=J\times 3$ holds stacked 3D coordinates of $J$ joints in a human/animal skeleton). We assume that each dataset is generated according to a set of \emph{discrete} latent states $\mathcal{S}_n=\{s_{n,t}\}_{t=1}^T$ and their corresponding low-dimensional \emph{continuous} temporal latent variables $Z_n = \{z_{n,t}\in \mathbb{R}^{K}\}_{t=1}^T$:
%, as follows:
\begin{align}
    X_n &\sim \text{Norm}(L_\theta[Z_n],\, \sigma^{X} I),\nonumber\\
    Z_n &\sim p_\theta(Z_n\,|\,\mathcal{S}_n),\nonumber\\
    \mathcal{S}_n &\sim p_\theta(\mathcal{S}),
    \label{eqn:pgen}
\end{align}
where, $L_\theta[Z_n]$ is a linear operator that projects local continuous latents $Z_n$ into the observation space (i.e., $L_\theta[Z_n] = Z_n^\top W_\theta$, where $W_\theta$ is a projection matrix), $\sigma^X$ denotes observation noise, $p_\theta(Z_n|\mathcal{S}_n)$ is a deep switching vector autoregressive (VAR) prior over $Z_n$ which is parameterized by neural networks, and $p_\theta(\mathcal{S})$ is a generative Markovian prior over local discrete latents $\mathcal{S}_n$. We collectively denote generative model parameters by $\theta$. The graphical representation for our proposed generative model is depicted in \figref{graph}.    

\subsection{Discrete Markovian Prior $\boldsymbol{p_\theta(\mathcal{S})}$}
We assume that for each dataset the temporal generative process resides at a specific state $s_t$ at time $t$ (out of $S$ possible states) which is determined according to a Markovian prior conditioned on its preceding continuous latent $z_{t-1}$. As such, the discrete latent states $\mathcal{S}_n=\{s_{n,t}\}_{t=1}^T$ are structured in a Markov chain as follows:
\begin{equation}
    p_\theta(s_t|s_{t-1}=s, z_{t-1}) = \text{Cat}\left(\sigma\left(\mathbf{\Phi}^s_{\theta}\,z_{t-1}\right)\right),
\end{equation}
where $\mathbf{\Phi}^s_\theta\in\mathbb{R}^{S\times K}$ is a state-specific transition matrix and $\sigma(\cdot)$ is a \texttt{softmax} function that ensures a valid $S$-dimensional probability vector. As noted in \cite{linderman2017bayesian}, conditioning the discrete states on their preceding continuous latents (in addition to their preceding discrete states) is desirable as it allows informed transitions. 

\subsection{Deep Switching VAR Prior $\boldsymbol{p_\theta(Z_n|\mathcal{S}_n)}$}
We assume that the low-dimensional dynamical latents $Z_n$ follow a nonlinear vector autoregressive Gaussian prior switched by their associated discrete states $\mathcal{S}_n$. This implies a Gaussian mixture distribution for the dynamical latent space:
\begin{align}
   p_\theta(z_{t}|z_{t-\ell},s_{t} = s) =
   \text{Norm}\Big(\mu_{\theta}^s(z_{t-\ell}),
   \sigma_{\theta}^s(z_{t-\ell})\Big),
\end{align}
where $s\in\{1,\cdots,S\}$ and $\ell$ denotes a lag set (e.g., $\ell=\{1,2\}$ for a second-order Markov model), and state-specific $\mu_{\theta}^s(\cdot)$ and $\sigma_{\theta}^s(\cdot)$ are parameterized by multilayer perceptrons (MLPs) (see \tabref{Arch}). In other words, we feed $z_{t-\ell}$ to a multi-head MLP for estimation of the Gaussian parameters, e.g.,
\begin{equation}
    \mu_{\theta}^s(z_{t-\ell}) = \sum_{l\in\ell} \text{MLP}_{\theta}^{s,l}(z_{t-l}).
    \label{eqn:sum}
\end{equation}

\subsection{Approximate Variational Inference}
%The graphical representation for our proposed generative model is depicted in \figref{graph}.
%\graphicalmodel

As the posterior probability for this model is intractable, we use approximate variational methods to learn the model parameters \citep{hoffman2013stochastic,ranganath2013adaptive}. These methods approximate the posterior of latents $p_\theta(\mathcal{S}, Z\,|\,X)$ with a variational distribution $q_\phi(\mathcal{S},Z)$ by maximizing the evidence lower bound (ELBO):
\begin{align}
    \mathcal{L}(\theta, \phi) =&\, \mathbb{E}_{q_\phi(\mathcal{S},Z)}\left[\log\frac{p_\theta(X, \mathcal{S}, Z)}{q_\phi(\mathcal{S},Z)}\right]\nonumber\\
    =&\log p_\theta(X) - \text{KL}\left(q_\phi(\mathcal{S},Z)\,\|\,p_\theta(\mathcal{S},Z|X)\right)
    \label{eqn:e}
\end{align}
By maximizing ELBO with respect to the parameters $\theta$, we learn a generative model that defines a distribution
over datasets $p_\theta(X)$. By maximizing ELBO over the parameters $\phi$, we perform Bayesian inference.

\subsubsection{Variational Distribution}
%\noindent\textbf{Variational Distribution:}
We assume a fully factorized variational distribution for the latents $\{\mathcal{S},Z\}$:
\begin{equation}
    q_\phi(\mathcal{S},Z) = \prod_{n=1}^N \prod_{t=1}^T
     q_\phi(s_{n,t}) q_\phi(z_{n,t}),
     \label{eqn:qinf}
\end{equation}
where $q_\phi(z_{n,t}) = \text{Norm}(\mu_\phi^{n,t}, \sigma_\phi^{n,t})$, and the categorical distributions $q_\phi(s_{n,t})$ are approximated with posteriors $p(s_{n,t}|z_{n,t})$, where $z_{n,t}\sim q_\phi(z_{n,t})$, to compensate information loss induced by the mean-field approximation:
\begin{align}
    q_\phi(s_{n,t}=s)&\simeq p(s_{n,t}=s|z_{n,t})\nonumber\\
    &=\frac{p_\theta(s_{n,t}=s)p_\theta(z_{n,t}|s_{n,t}=s)}{\sum_{s=1}^S p_\theta(s_{n,t}=s)p_\theta(z_{n,t}|s_{n,t}=s)}
\end{align}

\Arch
\subsubsection{ELBO Derivation}
%\noindent\textbf{ELBO Derivation:}
We can derive ELBO by plugging in the generative $p_\theta(X, \mathcal{S}, Z)$ and variational $q_\phi(\mathcal{S},Z)$ distributions from \eqnref{pgen} and \eqnref{qinf} respectively into \eqnref{e} (subscript and summation over $n$ are dropped for brevity):
\begin{align}
-\mathcal{L}(&\theta,\phi)=\nonumber\\
&\mathbb{E}_{q_\phi(Z)}\Big[\big\|X - L_\theta[Z]\big\|_\text{\tiny F}^2\Big]+\nonumber\\
&\sum_{t}\mathbb{E}_{q_\phi(s_{t-1},z_{t-1})}\Big[\text{KL}\big(q_\phi(s_{t})||p_\theta(s_{t}|s_{t-1},z_{t-1})\big)\Big]+\nonumber\\
&\sum_t\mathbb{E}_{q_\phi(s_t, z_{t-\ell})} \Big[\text{KL}\big(q_\phi(z_{t})\|
     p_\theta\big(z_{t}|z_{t-\ell},s_{t})\big)\Big],\nonumber
\end{align}
where the three terms correspond to reconstruction loss, discrete latent loss, and continuous latent loss, respectively.
%where the three terms correspond to reconstruction loss $\mathcal{L}^\text{rec}$, discrete latent loss $\mathcal{L}_{t}^{\mathcal{S}}$ and continuous latent loss $\mathcal{L}_{t}^{Z}$, respectively.
%\boldsymbol{\mathcal{L}^\textbf{rec}}
%\boldsymbol{\mathcal{L}_{t}^{\mathcal{S}}}
%\boldsymbol{\mathcal{L}_{t}^{Z}}
%\pendulum 
\section{Implementation Details}
We implemented our model in PyTorch v1.3 \citep{paszke2017automatic} and run our experiments on an Intel Core i7 CPU@3.7GHz with 8 GB RAM. Our model has $O($NKT$)$ variational and $O($S$|\ell|$K$^2)$ temporal generative parameters. We employed Adam optimizer \citep{kingma2014adam} with \texttt{lr=0.01} and estimated the gradients of ELBO using reparameterization trick, \citep{kingma2014auto}, for the continuous latent $Z$. The expectations over discrete latent $\mathcal{S}$ are easily handled by summing over all possible states. We trained our model for $1000$ epochs and each epoch took from $30$ to $500$ milliseconds in different experiments.

\section{Performance Measure}
%\noindent\textbf{Performance Measure:}
In order to quantify the performance of our dynamical generative model, we compute its temporal predictive error on a test set. To this end, we predict the next time point on a test set using the generative model learned on our train set: $\hat{x}_{t+1}=L_\theta(\hat{z}_{t+1})$, where $\hat{z}_{t+1}\sim p(\hat{z}_{t+1}|{z}_{t+1-\ell}, \hat{s}_{t+1})$ and $\hat{s}_{t+1}\sim p(\hat{s}_{t+1}|s_t,z_t)$. We then run inference on $x_{t+1}$, the actual observation at $t+1$, to obtain $z_{t+1}$ and $s_{t+1}$, and add them to the historical data for prediction of the next time point $\hat{x}_{t+2}$ in the same way. We repeat these steps to make predictions in a rolling manner across a test set and report their normalized root-mean-square error (NRMSE\%). We keep the generative model fixed during the entire prediction. Note that the test set prediction NRMSE\% is related to the expected \emph{negative test-set log-likelihood} for our case of Gaussian distributions (with a multiplicative/additive constant).

\section{Experimental Results}

We evaluated the performance of our proposed generative model in dynamical modeling and segmentation of motion data on a simulated physical system (pendulum system), three human motion data (Salsa dancing, walking and golf) and an animal motion data (bat flight). The generative and predictive performance of our model are summarized in \figref{pendulum}, \figref{salsa_generative}, \figref{bat_generative}, and \tabref{ErrorTable}.
%\par

%\subsubsection{Baselines} 
%\pendulum
\noindent\textbf{Comparison Baselines --}
We assessed our model against two established Bayesian switching dynamical models, recurrent switching linear dynamical systems (rSLDS) \citep{nassar2019tree} and switching linear dynamical systems (SLDS) \citep{fox2009nonparametric}, a state-of-the-art dynamical matrix factorization method, Bayesian temporal matrix factorization (BTMF) \citep{sun2019bayesian}, which models higher-order linear dependencies, a state-of-the-art deep
state-space model, recurrent Kalman networks (RKN) \citep{becker2019recurrent}, which employs first-order nonlinear transitions, and a deep neural network forecasting method, long- and short-term time-series network (LSTNet) \citep{lai2018modeling}, which employs vector auto-regression, throughout the experiments.

%The generative and predictive performance of our model are summarized in \figref{pendulum}, \figref{salsa_generative}, \figref{bat_generative}, and \tabref{ErrorTable} respectively.
\pendulum 
%\salsastates
%\salsagenerative
%\batstates
%\batgenerative
%\salsastates
%\salsagenerative
\subsection{Single Pendulum System}
A simple pendulum system shares appealing similarities with a joint-angle representation, and its motion is governed by a second-order nonlinear differential equation which makes it an interesting experiment for the purpose of this paper:
\begin{equation}
    \ddot{\theta} + g \sin(\theta) = 0
\end{equation}
where $\theta$ is the deflection angle of the pendulum and $g$ is the gravitational acceleration. We simulated this pendulum system for $T=400$ time points and recorded its 2D coordinates. We trained our model (and the baselines accordingly) with lag set $\ell=\{1,2\}$, two states $S=2$ and latent dimension $K=2$ on half of this dataset and kept the second half for test. Our model decomposed this motion data into two states: clockwise and anticlockwise rotation as depicted in \figref{salsa_states}. We have also visualized the dynamical trajectory of these two states using our generative model. 
Our model predicted the test set with NRMSE of $5.97\%$ (surpassing all the baselines) while rSLDS predicted with $25.58\%$ error. This is expected as the first-order linear transitions in rSLDS are not able to effectively model the higher-order and nonlinear dependencies in this model. Fitting our model with a single lag $\ell=\{1\}$ increased the error to $7.54\%$. This is also anticipated as the pendulum equation contains second derivative of location (i.e., acceleration). As reported in \tabref{ErrorTable}, the linear baselines (rSLDS, SLDS and BTMF) fail to capture the nonlinear transitions and as a result their predictive performance degrades significantly, whereas nonlinear models (Ours, RKN, LSTNet) perform better.
\subsection{Salsa Dance Dataset}

%We evaluate the performance of our method in dynamical modeling and segmentation of 3D human pose data. 
This dataset from CMU MoCap\footnote{http://mocap.cs.cmu.edu/} contains 3D coordinates of $19$ joints recorded for $T=200-571$ time points (every $100$ milliseconds) for 15 trials of Salsa dancing. We kept one trial for test, and only used the woman dancer data. We organized this dataset into a tensor of size $15\times T \times (19\times 3)$. We fit our model (and the baselines accordingly) on this human motion data with $S=3$, $\ell=\{1,2\}$, and $K=10$. As depicted in \figref{salsa_states}, our model segmented the sequences into 3 modes of motion which can be interpreted as: clockwise (CW) turn, anticlockwise (ACW) turn, and twirling motion. We have also computed the dynamical trajectory of each state purely from our learned generative model, and visualized that in \figref{salsa_generative}. To this end, we just feed the first two time points of the test set to our generative model (since we are using two lags), fix the state, and predict the next $100$ time points sequentially using our dynamical generative model (separately for $s=\{1,2,3\}$): 
\begin{align}
&\hat{z}_{t+1}\sim p_\theta(\hat{z}_{t+1}|\hat{z}_{t+1-\ell}, s_{t+1}=s)\nonumber\\
&\hat{x}_{t+1}= L_\theta(\hat{z}_{t+1})\quad \text{for}\;t=\{1,\cdots,100\}\nonumber
\end{align}
This gives us the state-specific dynamical trajectories visualized in \figref{salsa_generative} which perfectly follow our interpretation of each state. As reported in \tabref{ErrorTable}, our model predicted the test set with NRMSE of $6.74\%$ outperforming all the baselines. We have visualized test set predictions along with their uncertainty intervals for two sample joints in \figref{predictions} (a). We have rendered the test set and the generated dynamical trajectories for each state on a rigged 3D model of a salsa dancer in \texttt{Blender software}, \citep{blender}, and included their videos in our supplementary submission.       
\salsastates
\salsagenerative
\batstates
%\ErrorTable

\subsection{Bat Flight Dataset \citep{bergou2015falling}}

This dataset includes 3D coordinates of 34 joints on a bat body recorded over time for $T=166-436$ time points (every 33 milliseconds) during a landing/falling maneuver for 10 experimental runs with 32.55\% missing values as joint markers are frequently occluded during the flight. We held two runs out for test. We learned our model (and the baselines accordingly) on this data with $S=2$, $\ell=\{1,2\}$, and $K=5$. As depicted in \figref{bat_states}, our model appears to have parsed the bat flight motion into two modes of ``extending`` (i.e., stretching the wings), and ``flexing`` (folding the wings) which together constitute the ``flapping'' flight in birds. Similar to the Salsa dancing data, we predicted the dynamical trajectory of each state sequentially for $100$ time points (entirely from our learned dynamical model), and visualized that in \figref{bat_generative}. From this figure, the dynamical trajectory of each state completely support our interpretation of the states. As reported in \tabref{ErrorTable}, our model predicted the test set with NRMSE of $7.69\%$ outperforming all the baselines. We have also visualized test set predictions along with their uncertainty intervals for two sample joints in \figref{predictions} (b). Note that our model fills in missing values for this dataset. 
%As with salsa data, 
We have rendered the test set and the generated dynamical trajectories for each state on a rigged 3D model of a bat in \texttt{Blender software}, and included their videos in our supplementary submission.
\batgenerative
\walkstates
\subsection{Walking Dataset}

This is another dataset from CMU MoCap which contains 3D motion capture recordings from a subject for $34$ trials of walking/running. We kept two trials for test and trained our model on the rest with $S=2$, $\ell=\{1,2\}$, and $K=5$. The dynamical segmentation results for the two trials in the test set are visualized in \figref{walk_states}, and show that our model has parsed these locomotion sequences into two phases of ``right-leg swing'' and ``left-leg swing'' (encoded by blue and red colors, respectively), which are the familiar components during a bipedal gait cycle. As reported in \tabref{ErrorTable}, our model predicted the test set with NRMSE of $8.01\%$, significantly outperforming all of the baselines. We have visualized test set predictions along with their uncertainty intervals for two sample joints in \figref{predictions} (d).
\subsection{Golf Dataset}
This dataset is from the ``physical activities and sports'' part of CMU MoCap and includes $30$ trials of motion recordings from a subject while performing typical actions in a golf game including swing, placing ball, and picking up ball. We kept four trials for test, including two trials of swing and two trials of placing/picking up ball, and trained our model on the rest with $S=4$, $\ell=\{1,2\}$, and $K=10$. The dynamical segmentation results for the four trials in the test set are visualized in \figref{golf_states}. For swing trials, as pictured in the left side of \figref{golf_states}, our model divided swinging motion into major phases of ``backswing'' (heave+tip point), encoded by green+red colors, and ``downswing'' (fall+release), encoded by blue+yellow colors. Similarly, for the ball placement/pick-up trials in the right side of \figref{golf_states}, our model split body motion into distinct phases of ``bending down'', encoded by blue+yellow colors, and ``standing up'', encoded by green+red colors. As reported in \tabref{ErrorTable}, our model predicted the test set with NRMSE of $10.57\%$, surpassing rSLDS, SLDS, RKN, and LSTNet while closely following BTMF. We have visualized test set predictions along with their uncertainty intervals for two sample joints in \figref{predictions} (c).

\golfstates
\ErrorTable
\preds
\section{Ablation Study}
We conducted an ablation study to evaluate the impact of switching feature, $\mathcal{S}$, and temporal lags, $\ell$, in our model in terms of prediction accuracy. To this end, we executed a version of our model without the switching feature, denoted by \texttt{Ours w/o$\,$switch}, and a version with first-order temporal lag, denoted by \texttt{Ours w/$\,\ell=\{1\}$}, and applied them on our experimental datasets. The results of test set predictions for these model variants are compared with the original model in \tabref{AblationTable}. It is clear from the results of this table that both the switching feature (i.e., dynamical modes) and higher-order temporal modeling have consistently enhanced prediction accuracy of our model in all the experiments.   

\ablTable

\section{Conclusion}
We proposed a deep switching dynamical model for dynamical analysis of 3D motion data. Our model was able to uncover interpretable states in the low-dimensional dynamical generative model of the data. We parameterized these low-level temporal generative models with regard to a switching deep vector autoregressive (VAR) prior to enable multimodal and higher-order dynamical estimation.  Our segmentation, generative and predictive results on one simulated physical system and four real motion data demonstrated the superior performance of the proposed model in comparison with the state-of-the-art methods.

%\begin{acknowledgements}
%\end{acknowledgements}


\bibliographystyle{spbasic}      % basic style, author-year citations
\bibliography{ref}   % name your BibTeX data base

\end{document}

%% file: figs.tex
\newcommand{\graphicalmodel}{
\begin{figure}
 \centering
 \includegraphics[width=1\linewidth]{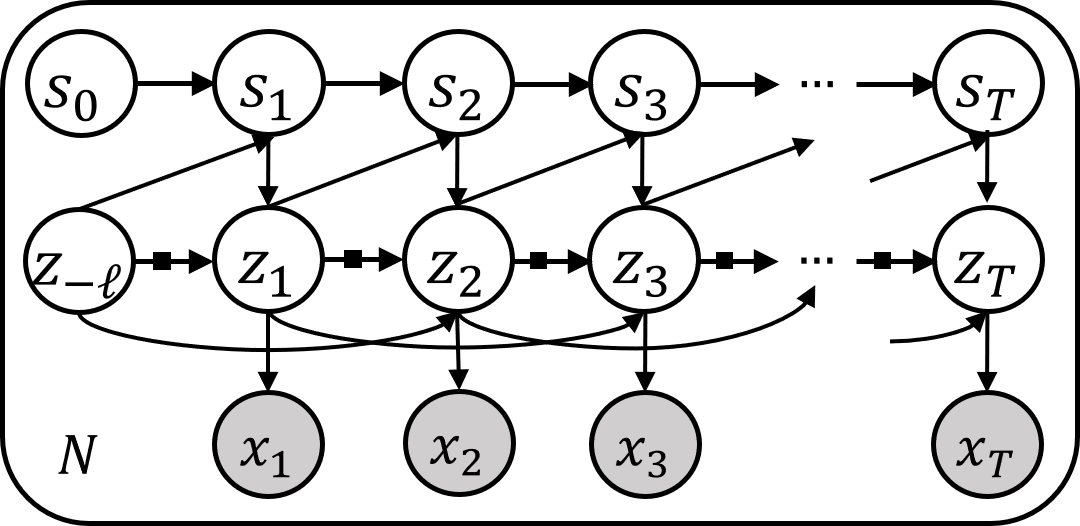}
 \caption{Graphical representation for our deep switching dynamical model given $N$ motion datasets. The low-dimensional temporal latents $Z_n=\{z_{n,t}\}_{t=1}^{T}$ are generated with regards to a nonlinear vector autoregressive (VAR) prior switched by their associated discrete states $\mathcal{S}_n=\{s_{n,t}\}_{t=1}^{T}$. The discrete states themselves are determined according to a Markovian prior conditioned on their preceding continuous latents, i.e, $z_{t-1}$. The solid black squares represent nonlinear functions. Gray-shaded circles represent observations.}
 \label{fig:graph}
\end{figure}
}

\newcommand{\pendulum}{
\begin{figure*}
 \centering
 \includegraphics[width=1\linewidth]{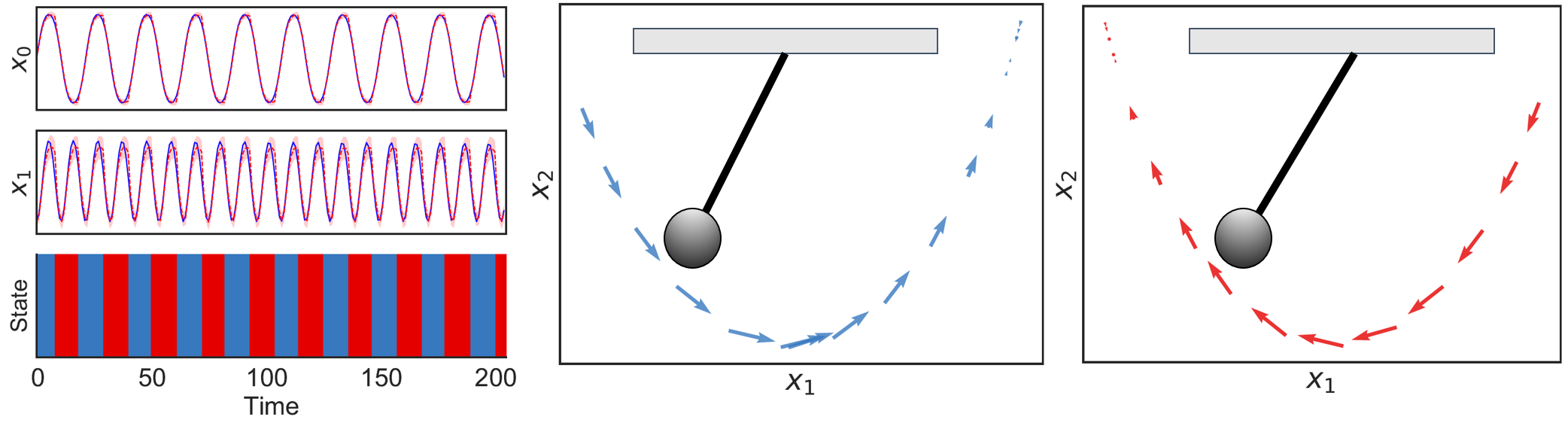}
 \caption{Inferred states (left) and dynamical trajectories (right) in a pendulum system. The motion in a pendulum system is governed by a second-order nonlinear differential equation. Our model decomposed this motion data into two states: clockwise and anticlockwise rotation. The dynamical trajectory of these two states are visualized using our learned generative model. Our model with lag set $\ell=\{1,2\}$ predicted the test set with NRMSE of $5.97\%$ significantly outperforming the baselines. This is expected as the first-order and/or linear transitions in the baselines are not able to effectively model the higher-order and nonlinear dependencies in this system. The true (blue) and predicted (red) 2D coordinates (left-top) are also shown. Red-shaded regions show prediction uncertainty.}
 \label{fig:pendulum}
\end{figure*}
}

\newcommand{\salsastates}{
\begin{figure*}[t]
 \centering
 \includegraphics[width=1.0\linewidth]{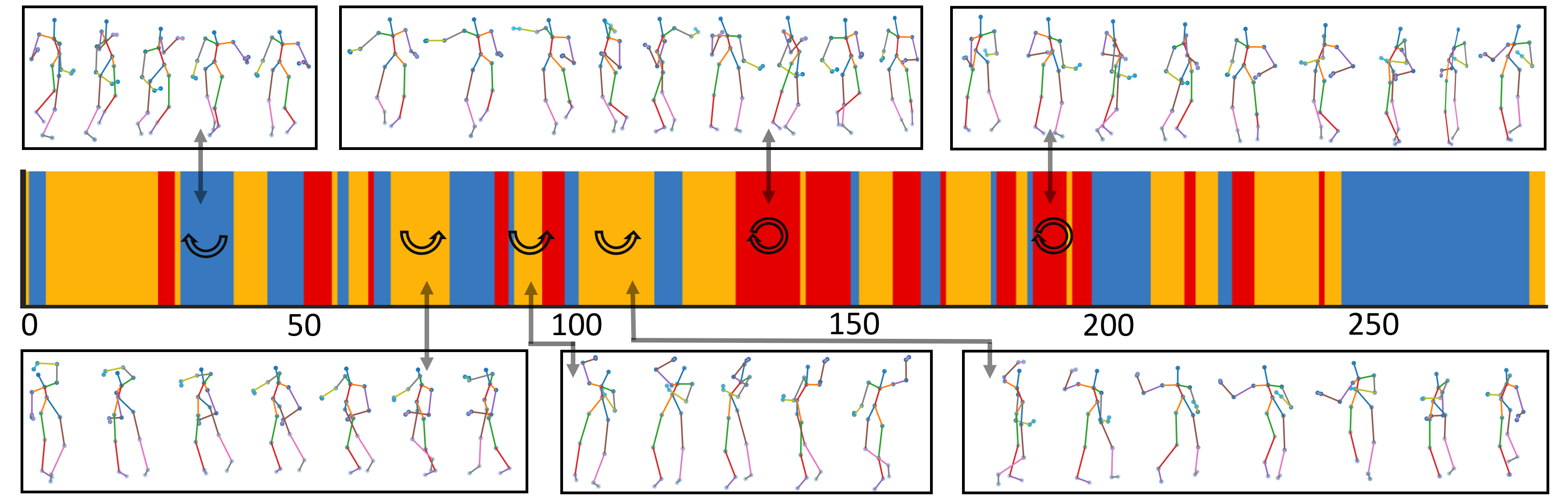}
 \caption{Inferred states in the test set of salsa dance data. Our model segmented the data into three modes of motion which can be interpreted as: clockwise (CW) turn, anticlockwise (ACW) turn, and twirling motion.}
 \label{fig:salsa_states}
\end{figure*}
}

\newcommand{\salsagenerative}{
\begin{figure}[t]
 \centering
 \includegraphics[width=1\linewidth]{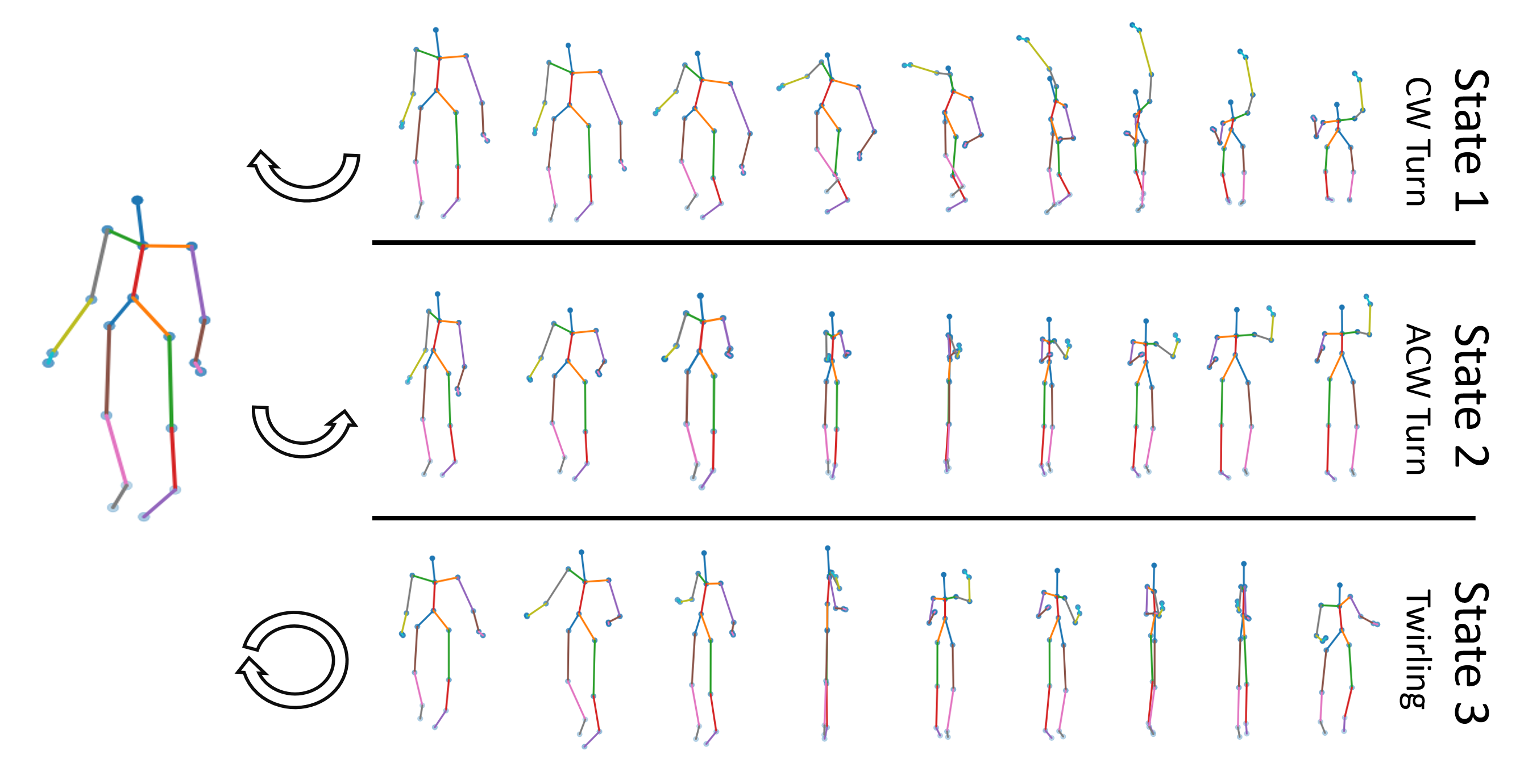}
 \caption{Dynamical trajectories for each state in the salsa dance data (clockwise (CW) turn, anticlockwise (ACW) turn, and twirling motion). We computed the dynamical trajectory of each state purely from the learned generative model. To this end, we just feed the first two frames of the test set to our generative model, fix the state, and predict the next $100$ time points sequentially using our dynamical generative model: $\hat{z}_{t+1}\sim p_\theta(\hat{z}_{t+1}|\hat{z}_{t+1-\ell}, s_{t+1}=s)$ and $\hat{x}_{t+1}= L_\theta(\hat{z}_{t+1})$ for $t=\{1,\cdots,100\}$. The predicted dynamical trajectories perfectly follow our interpretation of each state.}
 \label{fig:salsa_generative}
\end{figure}
}

\newcommand{\batstates}{
\begin{figure*}[t]
 \centering
 \includegraphics[width=.95\linewidth]{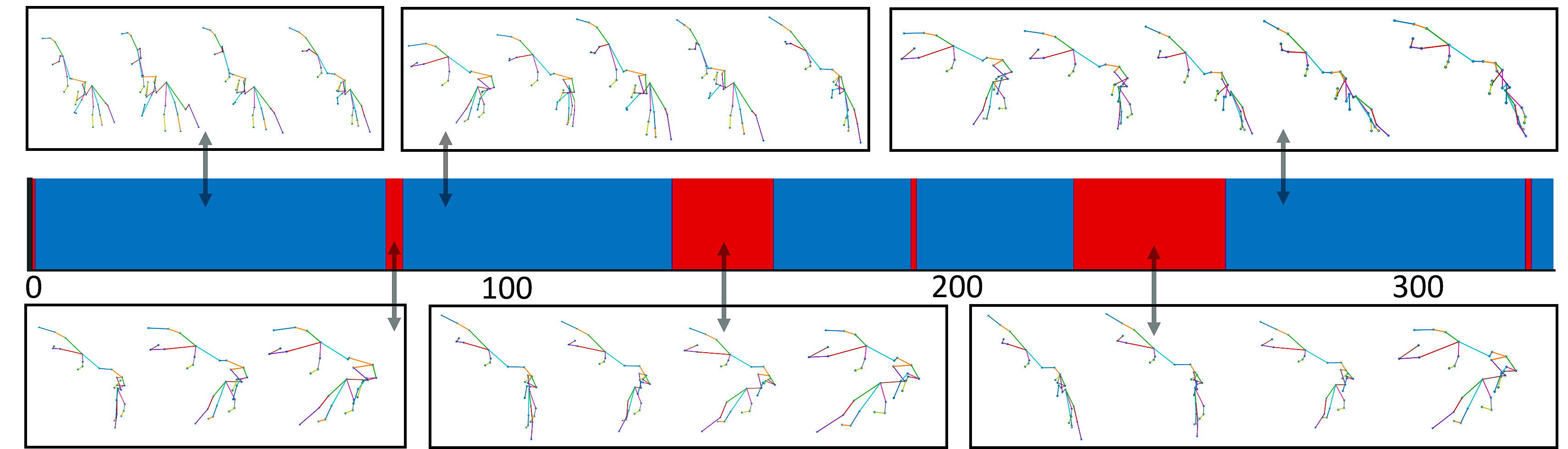}
 \caption{Inferred states in the test set of bat flight data. Our model has parsed the bat motion into two modes of ``extending`` and ``flexing`` which together constitutes the ``flapping`` flight in birds.}
 \label{fig:bat_states}
\end{figure*}
}

\newcommand{\batgenerative}{
\begin{figure}[t]
 \centering
 \includegraphics[width=1\linewidth]{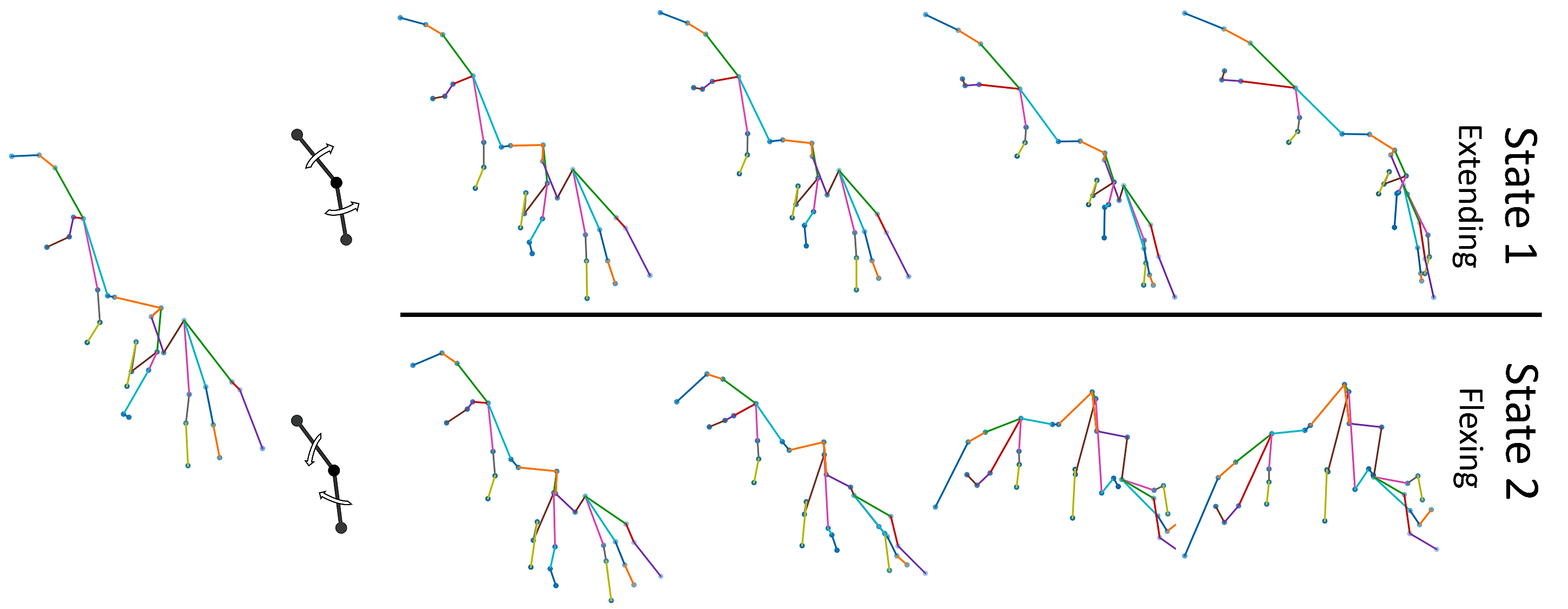}
 \caption{Dynamical trajectories for each state in bat flight data. We predicted the dynamical trajectory of each state sequentially for $100$ time points entirely from our learned dynamical generative model (given the initial frame(s) of the test set). The state-specific dynamical trajectories completely support our interpretation of the states.
 }
 \label{fig:bat_generative}
\end{figure}
}

\newcommand{\preds}{
\begin{figure*}
 \centering
 \includegraphics[width=1\linewidth]{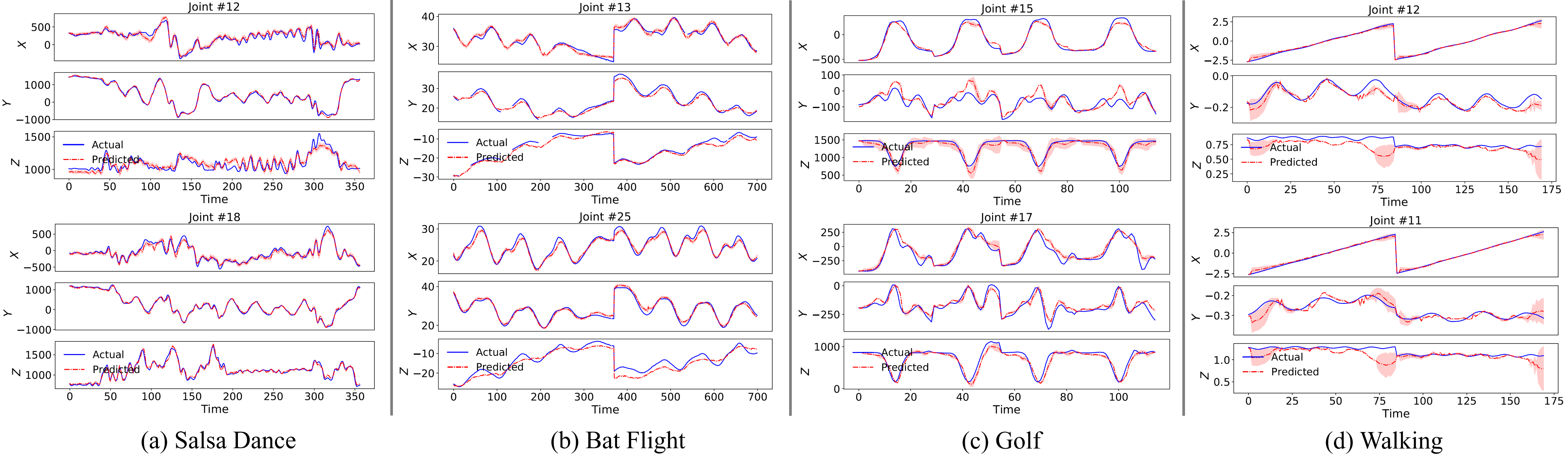}
 \caption{Visualizations of test set predictions for two sample joints for each of the salsa dance, bat flight, golf and walking datasets. Note that our model fills in missing values in the bat flight dataset. Red-shaded regions correspond to prediction uncertainty.}
 \label{fig:predictions}
\end{figure*}
}

\newcommand{\golfstates}{
\begin{figure*}[t]
 \centering
 \includegraphics[width=1\linewidth]{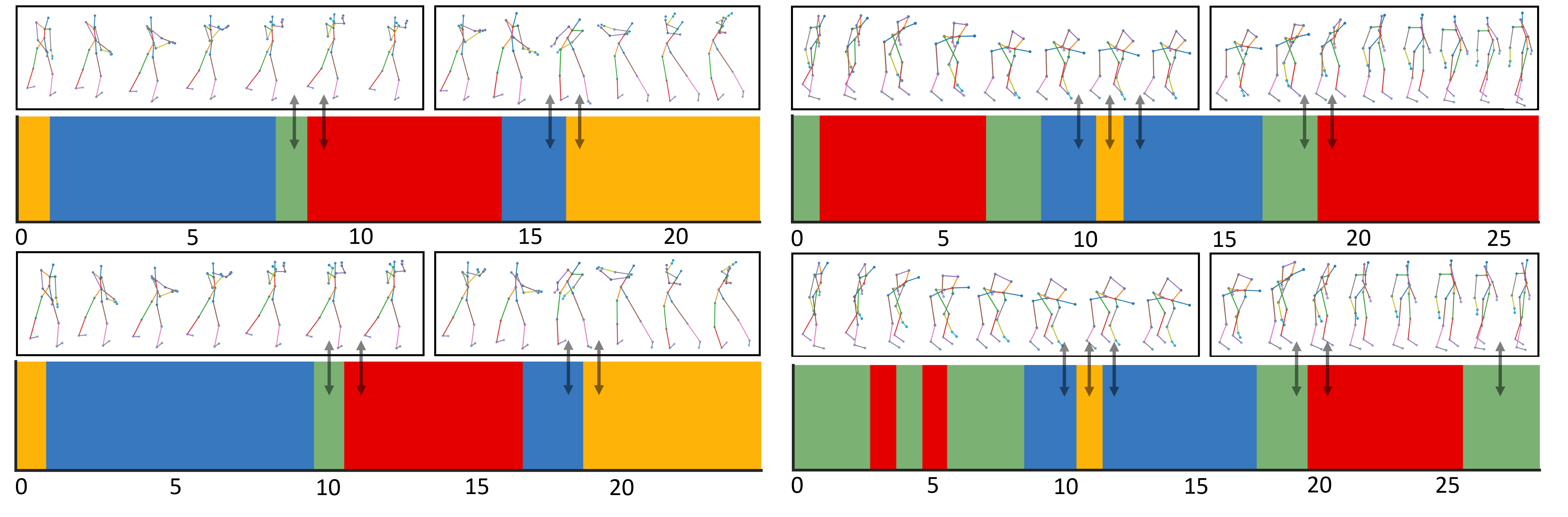}
 \caption{Inferred dynamical segmentation for the four trials in the test set of golf dataset, including two trials of swing (left) and two trials of placing/picking up ball (right). \textbf{Left:} For swing trials, our model divided the swinging motion into major phases of ``backswing'' (heave+tip point), encoded by the green+red colors, and ``downswing'' (fall+release), encoded by the blue+yellow colors. \textbf{Right:} For the ball placement/pick-up trials, our model split body motion into distinct phases of ``bending down'', encoded by the blue+yellow colors, and ``standing up'', encoded by the green+red colors.}
 \label{fig:golf_states}
\end{figure*}
}

\newcommand{\walkstates}{
\begin{figure*}[b]
 \centering
 \includegraphics[width=1\linewidth]{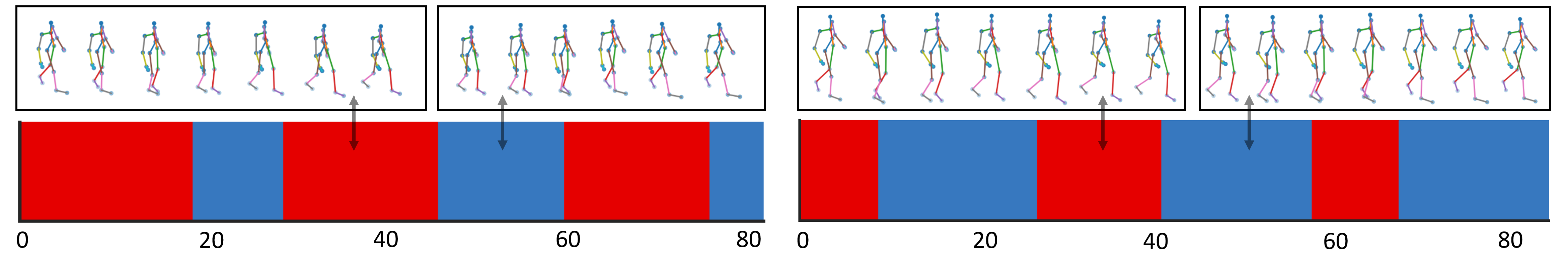}
 \caption{Inferred dynamical states for the two trials in the test set of walking dataset. Our model has parsed these locomotion sequences into two phases of ``right-leg swing'', encoded by the blue color, and ``left-leg swing'' encoded by the red color, which are the familiar components during a bipedal gait cycle.}
 \label{fig:walk_states}
\end{figure*}
}

%% file: tabs.tex
\newcommand{\Arch}{
\begin{table}
\centering
\caption{Network architecture for MLPs parameterizing the Gaussian distribution $p_\theta\big(z_t | z_{t-\ell}, s_t = s\big)$. We defined a fully connected (FC) layer for each state $s\in \{1, . . . , S\}$, and lag $l \in \ell$ as FC$_{s,l}$, the output of which are summed in the succeeding layer according to \eqnref{sum}.}
\vspace{-.2cm}
{\footnotesize
\setlength\tabcolsep{2pt}
\begin{tabular}{|c|}
\toprule
 $\boldsymbol{p_\theta\big(z_t | z_{t-\ell}, s_t = s\big)}$\\
  \toprule
 
 $\textbf{Input:}\;\boldsymbol{{z_{t-\ell}}}\boldsymbol{\in\mathbb{R}^{|\ell|\times K}}$\\
 
 \toprule
 
 \text{$\big[\text{FC}_{\theta}^{s,l}$ $K\times K\big]_{l=1}^{|\ell|}$ $\text{(PReLU)}\rightarrow{(*)}$}\\
 
 \text{$\sum_{l}{(*)}\rightarrow\text{FC}_{\theta}^s$ $K\times K$}
 \\
 $\textbf{Output:}\;\boldsymbol{\mu^{s}_{\theta}\in \mathbb{R}^{K}}$
 \\
 \text{(PReLU)} \text{$\text{FC}_{\theta}^s$ $K\times K$}
 \\
 $\textbf{Output:}\;\boldsymbol{\textbf{log}\Sigma^s_{\theta}\in \mathbb{R}^{K}}$
 \\
 \bottomrule
 \end{tabular}
}
 \label{tbl:Arch}
\end{table}
}

\newcommand{\ErrorTable}{
\begin{table}[b]
\caption{Comparison of prediction error (NRMSE\%) on the test sets of pendulum, salsa dance, bat flight, walking and golf datasets. Our model outperforms the baselines.}
\centering
{\small
\setlength\tabcolsep{2pt}
\begin{tabular}{l|c|c|c|c|c|c|c|}
\toprule
\backslashbox{Dataset}{Model} 
& \texttt{Ours}
& rSLDS
& SLDS
& BTMF
& RKN
& LSTNet
\\
\midrule
Salsa Dance 
& $\mathbf{6.74}$
& $8.78$
& $8.68$
& $7.75$
& $8.86$
& $7.91$
\\
Bat Flight
& $\mathbf{7.69}$
& $9.93$
& $10.69$
& $8.91$
& $17.81$
& $16.35$
\\
Golf
& $10.57$
& $10.89$
& $13.64$
& $\mathbf{9.43}$
& $12.60$
& $17.87$
\\
Walking
& $\mathbf{8.01}$
& $21.81$
& $22.59$
& $31.39$
& $23.56$
& $14.77$
\\
Pendulum
& $\mathbf{5.95}$
& $25.58$
& $29.02$
& $21.14$
& $8.77$
& $8.62$
\\
\bottomrule
\multicolumn{7}{l}{\scriptsize Best results are highlighted in bold fonts.}
\end{tabular}
}
\label{tbl:ErrorTable}
\end{table}
}

\newcommand{\ablTable}{
\begin{table}[b]
\caption{Comparison of our proposed model with its two ablated variants in terms of prediction error (NRMSE\%) on the test sets of our datasets. Both the switching feature and higher-order temporal modeling have consistently enhanced prediction accuracy of our model in all the experiments.}
\centering
{\small
\setlength\tabcolsep{2pt}
\begin{tabular}{l|c|c|c|}
\toprule
\backslashbox{Dataset}{Model}
& \texttt{Ours}
& \texttt{Ours w/o$\,$switch}
& \texttt{Ours w/$\,\ell=\{1\}$}
\\
\midrule
Salsa Dance 
& $\mathbf{6.74}$
& $8.73$
& $8.76$
\\
Bat Flight
& $\mathbf{7.69}$
& $8.34$
& $8.72$
\\
Golf
& $\mathbf{10.57}$
& $11.23$
& $11.79$
\\
Walking
& $\mathbf{8.01}$
& $9.85$
& $9.88$
\\
Pendulum
& $\mathbf{5.95}$
& $6.91$
& $7.54$
\\
\bottomrule
\multicolumn{4}{l}{\scriptsize Best results are highlighted in bold fonts.}
\end{tabular}
}
\label{tbl:AblationTable}
\end{table}
}